\documentclass[letterpaper, 10 pt, conference]{ieeeconf}  

\IEEEoverridecommandlockouts                              

\usepackage{graphics} 
\usepackage{epsfig} 
\usepackage{mathptmx} 
\usepackage{times} 
\usepackage{amsmath} 
\usepackage{amssymb}  

\title{\LARGE \bf
Analyzing the Influence of Dataset Composition for Emotion Recognition
}

\author{ Alexander Sutherland$^{1}$, Sven Magg$^{1}$, Cornelius Weber$^{1}$, Stefan Wermter$^{1}$ \\
\thanks{$^{1}$Knowledge Technology, Department of Informatics, University of Hamburg, Germany, \{{\tt\small sutherland, magg, weber, wermter\}@informatik.uni-hamburg.de}}}

\begin{document}

\maketitle
\thispagestyle{empty}
\pagestyle{empty}

\begin{abstract}

Recognizing emotions from text in multimodal architectures has yielded promising results, surpassing video and audio modalities under certain circumstances. However, the method by which multimodal data is collected can be significant for recognizing emotional features in language. In this paper, we address the influence data collection methodology has on two multimodal emotion recognition datasets, the IEMOCAP dataset and the OMG-Emotion Behavior dataset, by analyzing textual dataset compositions and emotion recognition accuracy. Experiments with the full IEMOCAP dataset indicate that the composition negatively influences generalization performance when compared to the OMG-Emotion Behavior dataset. We conclude by discussing the impact this may have on HRI experiments.
\end{abstract}

\section{Introduction}




One of the key components of affective computing is the ability to recognize emotion features in humans. Numerous approaches focus on combining features from multiple input modalities \cite{Poria2017review}. The most common modalities are centred on features collected from video and audio stimuli since these features are the easiest to collect and label. Recent experiments on multimodal emotion recognition have made prolific use of features of language and text. Poria et al. \cite{Poria2016covolutional} in particular have shown impressive results using CNNs when combining text with other modalities on experimental data, however, little attention has been spent on how and why textual features are able to perform so well in a multimodal context when compared to visual and acoustic features.

We address this issue by analyzing how the composition of the IEMOCAP dataset influences neuron activation during classification and show how current usage of the IEMOCAP manifests in visible over-fitting. Experimental results indicate that a network pretrained on the IEMOCAP barely performs as well as a network pretrained on the OMG-Emotion Behavior dataset \cite{Barros2018Omg} when attempting to classify the other dataset, in spite of being a larger dataset.

\section{Dataset Descriptions}

We make use of the IEMOCAP dataset \cite{Busso2008IEMOCAP} and the OMG Emotion-Behavior dataset \cite{Barros2018Omg}. The IEMOCAP dataset consists of videos containing either scripted or improvised utterances. Videos are recorded in different sessions, with different actors in each session. We merge the ``\textit{Happy}'' and ``\textit{Excited}'' data samples in the IEMOCAP in order to be able to emulate previous results from Poria et al. \cite{Poria2016covolutional} and to even out class balance. We use 5616 transcribed labelled utterances from the dataset with 1117 \textit{Angry} samples, 1644 \textit{Happy} and \textit{Excited} samples, 1753 \textit{Neutral} samples, and 1102 \textit{Sad} samples. The set of emotions chosen for this task is based on the data used in previous experiments on the same task \cite{Poria2016covolutional}. 



The OMG Emotion-Behavior dataset \cite{Barros2018Omg} consists of videos labelled through crowd-sourcing. Each video is annotated within the context of a longer clip, with both categorical and continuous emotion labels. Categorical labels include \textit{Anger}, \textit{Neutral}, \textit{Happy}, \textit{Sad}, \textit{Surprise}, \textit{Fear}, and \textit{Disgust}. We use 4656 samples consisting of 639 Sad, 665 Anger, 1794 Neutral, and 1558 Happy samples to compare a network's performance on the IEMOCAP for the same emotions. Videos were selected using a web-crawler that searched for Youtube videos, with keywords such as ``monologue'' and ``acting'', with each video being split into utterances. 

\section{Experimental Dataset Analyses}

This experiment determines the influence of the IEMOCAP dataset construction method on emotion recognition from text. Previous works have found emotion features at both word \cite{Poria2016covolutional} and semantic levels \cite{Wu2006emotion}. We use the pretrained Google News Word Embedding \cite{Mikolov2013Efficient} and a semantic embedding of size 50 trained on frames, from SEMAFOR \cite{Schneider2010semafor}, unimodally and multimodally as input to a 1D-CNN architecture. The architecture has channels for each input with a dropout of 0.2, a temporal convolution layer with a kernel size of 3, a stride of 1, and a filter size of 150, a global max pooling layer, a penultimate fully connected layer of size 32 and a final softmax layer of size 4. Late fusion between inputs is performed via concatenating max pooling outputs before feeding them to the fully connected layer. Classification accuracies from 10-fold cross-validations with 8-1-1 splits can be seen in Table \ref{tab:IEMOCAPonly} for both datasets, and Table \ref{tab:pretrain} presents the results of dataset generalization accuracy.

\begin{table}
\caption{\label{tab:IEMOCAPonly} Mean 10-fold accuracy and standard deviation for emotion recognition on selected labels from the IEMOCAP, only IEMOCAP improvised samples, and OMG textual data using 1D-CNNS for word embeddings, semantics, and a fusion. }
\begin{center}
  \begin{tabular}{ | l | c | c | c |c |}
    \hline
     & IEMOCAP & IEMOCAP Improv. & OMG \\ \hline
    Words & 67.504 (1.436) & 62.738 (2.490) & 44.200 (1.377)  \\ \hline
    Semantics & 52.672 (2.306) & 47.330 (3.512) &  39.371 (1.337)  \\ \hline
    Fusion & 67.718 (2.235) & 63.075 (1.876) & 43.557 (2.165) \\
    \hline
  \end{tabular}
\end{center}
\end{table}

  

\begin{table}
\caption{\label{tab:pretrain} Mean 10-fold accuracy and standard deviation when pretraining the CNN on the dataset to the left of the arrow and testing on the right for text emotion recognition. Datasets used are the IEMOCAP, \textbf{I}, the improvised samples of the IEMOCAP, \textbf{I improv.}, and the OMG dataset, \textbf{O}.}
\begin{center}
  \begin{tabular}{ | l | c  | c | c |}
    \hline
      & \textbf{I} $\rightarrow$ \textbf{O}   & \textbf{I Improv.} $\rightarrow$ \textbf{O} & \textbf{O} $\rightarrow$ \textbf{I} \\ \hline
    Words & 34.487 (0.862)   &  35.195 (1.174) &  \textbf{35.589} (0.710)\\ \hline
    Semantics & 30.483 (1.152)  &  \textbf{33.251} (0.766) &  31.193 (1.059)  \\ \hline
    Fusion & 35.217 (1.636)   &  \textbf{35.612 }(1.023) &  35.185 (0.547)  \\
    \hline
  \end{tabular}
\end{center}
\end{table}


\section{Results \& Discussion}
In Table \ref{tab:IEMOCAPonly} we see the results of emotion classification accuracy for input feature combinations. Clearly word embeddings are the primary contributor to classification results and that the network attains higher accuracy on the IEMOCAP as opposed to the OMG. Table \ref{tab:pretrain} shows results of pretraining the network on one dataset and classifying samples from the other. We see that the full IEMOCAP performs equal to or worse than the OMG and IEMOCAP improvised samples when generalizing, despite the full IEMOCAP's high performance on its own test data. 

\begin{figure}
  \centering
      \includegraphics[width=0.5\textwidth]{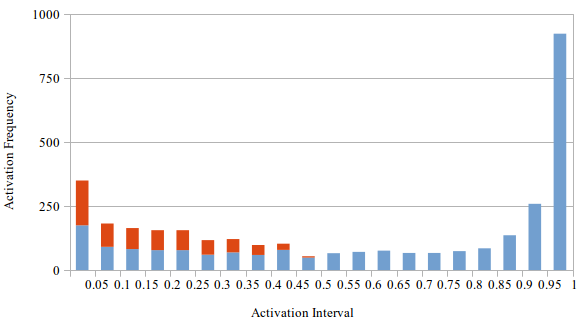}
  \caption{Scripted utterance softmax activation frequencies predicted by the utterance-level unimodal word model for the 2645 IEMOCAP scripted utterances. The blue bars indicate the number of target label activation levels occurring within particular activation brackets. Red bars indicate the number of incorrect classifications when a target label neuron activation is in a particular bracket. There is no red bar for activations above 0.45, as this entails that the classification was correct, i.e. above 0.5.}
   \label{scripthist}
\end{figure}

\begin{figure}

  \centering
      \includegraphics[width=0.5\textwidth]{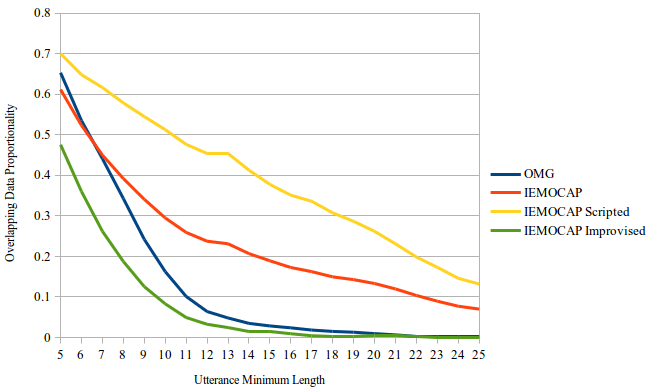}
  \caption{The proportion of data that overlaps with some other data point with regard to a certain number of shared words. The X-axis describes the number of words required for a sentence to be considered and the Y-axis describes what proportion of considered sentences overlap with some other sentence. The IEMOCAP scripted sessions suffer from the highest level of overlap while the IEMOCAP improvised sessions suffer from the least.}
   \label{datacomp}
\end{figure}

We believe this over-performance during testing but under-performance during generalization was due to the subset of scripted values in the IEMOCAP. To support this, in Figure \ref{scripthist}, we visualize the activation values attained when classifying scripted IEMOCAP utterances during the process of 10-fold cross-validation. We see abnormally high neuron activations in the 0.95 to 1 bracket, indicating a high certainty of correctness. The reason for this over-confidence is visualized in Figure \ref{datacomp}, where we show that scripted IEMOCAP utterances have a high number of data-points that partially overlap with others due to sentences being scripted and nearly identical. This is likely the reason for the textual modalities high performance in previously reported multimodal architectures that used the IEMOCAP dataset.



In summary, we have shown a reason for textual modalities high performance in previous works using the IEMOCAP in this manner. This over-fitting will also influence the understanding of natural language in applications, including HRI scenarios. Robot agents could place far greater salience in specific words than is desirable, such as the word ``beast'' in the IEMOCAP only occurring in angry data-points, leading to angry prone classifications. In future work, we suggest that researchers proceed with caution when selecting and applying datasets, as incorrect training procedures can lead to illogical and unreliable behaviour from models.





\section*{Acknowledgements}
This work has received funding from the European Union’s Horizon 2020 research and innovation programme under the Marie Sk{\l}odowska-Curie grant agreement No 721619 (SOCRATES).

\end{document}